\documentclass[lettersize,journal]{IEEEtran}
\usepackage{amsmath,amsfonts}
\usepackage{algorithmic}
\usepackage{algorithm}
\usepackage{array}
\usepackage[caption=false,font=normalsize,labelfont=sf,textfont=sf]{subfig}
\usepackage{textcomp}
\usepackage{stfloats}
\usepackage{booktabs}
\usepackage{url}
\usepackage{verbatim}
\usepackage{multirow}
\usepackage{colortbl}
\usepackage{graphicx}
\usepackage{cite}
\usepackage{xcolor}
\usepackage[normalem]{ulem}
\usepackage{soul}
\usepackage{hyperref}
\definecolor{mygray}{gray}{.9}

\begin{document}
	
	\title{Towards a Unified Transformer-based Framework for Scene Graph Generation  and Human-object Interaction Detection}
	
	\author{Tao He$^{1,2}$, Lianli Gao$^3$, Jingkuan Song$^3$, Yuan-Fang Li$^{2*}$ \thanks{$^*$Corresponding author.} \\ 
 $^1$Laboratory of Intelligent Collaborative Computing, University of Electronic Science and Technology of China
	$^2$Department of Data Science and AI, Faculty of Information Technology, Monash University \\
	$^3$Center for Future Media, University of Electronic Science and Technology of China\\
	{\tt\small tao.he01@hotmail.com,lianli.gao@uestc.edu.cn,jingkuan.song@gmail.com,yufang.li@monash.edu} 
	}
	
	\markboth{Journal of \LaTeX\ Class Files,~Vol.~14, No.~8, August~2021}%
	{Shell \MakeLowercase{\textit{et al.}}: A Sample Article Using IEEEtran.cls for IEEE Journals}
	

	\maketitle
	
	\begin{abstract}
	Scene graph generation (SGG) and human-object interaction (HOI) detection are two important visual tasks aiming at localising and recognising relationships between objects, and interactions between humans and objects, respectively. 
 Prevailing works treat these tasks as distinct tasks, leading to the development of task-specific models tailored to individual datasets. However, we posit that the presence of visual relationships can furnish crucial contextual and intricate relational cues that significantly augment the inference of human-object interactions. This motivates us to think if there is a natural intrinsic relationship between the two tasks, where scene graphs can serve as a  source for inferring human-object interactions. In light of this, we introduce SG2HOI+, a unified one-step model based on the Transformer architecture. Our approach employs two interactive hierarchical Transformers to seamlessly unify the tasks of SGG and HOI detection. Concretely, we initiate a relation Transformer tasked with generating relation triples from a suite of visual features. Subsequently, we employ another transformer-based decoder to predict human-object interactions based on the generated relation triples.
A comprehensive series of experiments conducted across established benchmark datasets including Visual Genome, V-COCO, and HICO-DET demonstrates the compelling performance of our SG2HOI+ model in comparison to prevalent one-stage SGG models. Remarkably, our approach achieves competitive performance when compared to state-of-the-art HOI methods. Additionally, we observe that our SG2HOI+  jointly trained on both SGG and HOI tasks in an end-to-end manner yields substantial improvements for both tasks compared to individualized training paradigms.

	\end{abstract}
	
	
\section{Introduction}

\IEEEPARstart{S}{cene}  graph generation (SGG) has {garnered substantial attention in recent years due to its potential in enhancing machine understanding of complex visual scenes~\cite{xu2017scene,lu2016visual,newell2017pixels,he2020learning,dai2017detecting}. A closely related task, human-object interaction (HOI) detection, is concerned with detecting the interactions occurring between humans and objects~\cite{gkioxari2018detecting,liao2020ppdm}. In both tasks, given an image, the objective is to localize all objects, including humans, and discern the predicate words \footnote{Throughout this paper, we consistently use the terms "relationship" and "interaction" interchangeably with "predicate words".} in the form of triples, denoted as $<$subject/human, predicate, object$>$. It is noteworthy that for HOI, the subject of any triple must be a human entity. In essence, a scene graph (SG) represents a general relation graph, whereas an HOI graph constitutes a human-centered subject graph, which can be perceived as a subgraph extracted from an SG.}


Although these two tasks share a common goal of detecting predicate words between two objects within an image, their development has largely been pursued independently. {This divergence can be attributed to two primary distinctions between them. Firstly, subjects in SGG can encompass a diverse range of entities such as humans, cars, and more, whereas in  HOI detection, subjects are confined to humans only. Secondly, the predicates of HOI are exclusively high-level interaction verbs that delineate the actions of humans, whereas a scene graph may encompass basic low-level relationship terms, such as locative prepositions (e.g. "on") as well as semantic actions (e.g. "play with").}



{
However, we believe that incorporating scene graphs into HOI detection offers notable advantages. Firstly, a scene graph encapsulates comprehensive visual relationships among objects in a structural graph, providing valuable contextual and high-level cues that augment scene understanding and subsequently lead to more accurate HOI inference. Secondly, the relationships in an SG can be either explicitly or implicitly transferred to interactions. For instance, given the relationships $<$hand, of, boy$>$ and $<$hand, holding, bread$>$, we
are more likely to infer that the boy's interaction related to bread is \textit{eating}. Thirdly, the simultaneous learning of both tasks within a unified model holds the potential to enhance data efficiency and benefits from multi-task training strategies, thereby refining the acquired representations for both tasks. In our preceding work, SG2HOI \cite{he2021exploiting}, we proposed the utilization of offline scene graphs generated by an off-the-shelf SGG model to enhance HOI detection. However, its notable limitation is the independent training of the SGG step and the HOI network. We posit that SG2HOI falls short in fully capitalizing on the benefits offered by multi-task learning and end-to-end training.}

Inspired by DETR~\cite{carion2020end}, an end-to-end object detection architecture based on the Transformer model~\cite{vaswani2017attention}, recent advancements in  SGG and HOI detection have introduced one-stage models that conceptualize relation or interaction detection as a problem of set prediction. In contrast, conventional two-step approaches in HOI and SGG~\cite{zellers2018neural,he2020learning,he2021exploiting,gkioxari2018detecting,qi2018learning} initially employ an off-the-shelf object detection network to localize potential objects and subsequently predict predicates based on pair-wise combinations of region proposals corresponding to subjects and objects. {One notable drawback of these methods is the division of optimization into two separate stages, which can lead to suboptimal solutions. This is partly due to the fact that manually crafted subject-object proposals often exhibit relatively low quality for predicate classification \cite{liao2020ppdm}. Additionally, such methods tend to incur redundant computation costs \cite{cong2022reltr,liao2022gen}.}

 
{Motivated by the aforementioned observations, we build upon our previous work \cite{he2021exploiting} and raise an intuitive research question: \textit{Can we develop a unified model that predicts both scene graphs and human-object interactions through a single, cohesively trained model?} In this endeavor, we present a unified \textit{one-stage} model that employs two sequential Transformer-based networks to jointly address SGG and HOI detection tasks. At its core, our approach formulates these tasks as consecutive set prediction problems. Concretely, we initially utilize a \emph{relation Transformer} to generate a  scene graph from a set of visual features. Subsequently, we employ an \emph{interaction Transformer} to predict HOIs by leveraging the generated scene graph, i.e., a set of relation triples. During the training process, both Transformers collaborate to  produce scene graphs and predict human-object interactions in an integrated and end-to-end manner.
}

A major challenge for one-stage models lies in the effective integration of semantic and spatial cues for relation and interaction classification~\cite{kim2020uniondet,wang2020learning,zou2021end}. 
This arises from the fact that one-stage models lack the access to object information prior to predicate classification, a capability enjoyed by two-stage models~\cite{zellers2018neural,tang2019learning,gao2018ican}. To tackle this challenge, Iftekhar \textit{et al.} \cite{iftekhar2022look} introduced a support feature generator designed to integrate semantic and spatial knowledge for HOI prediction.  However, their approach separated the training of spatial and semantic features using two distinct networks. We assert that this independent training strategy could potentially yield less robust representations, primarily due to the segregated training modules employed.


To overcome this challenge, drawing inspiration from recent advancements in image segmentation~\cite{minaee2021image}, we introduce a bounding-box-based image segmentation, offering a holistic semantic-spatial feature. By concatenating this semantic-spatial representation with visual features, we feed them into subsequent transformers. {This enables the joint learning of visual, semantic, and spatial information, eliminating the need for separate and manual embeddings as seen in prior works~\cite{zellers2018neural,he2021exploiting,iftekhar2022look}. For the task of SGG, following the mainstream one-stage models \cite{iftekhar2022look,cong2022reltr}  we leverage a Transformer-based relation decoder that takes as input the encoded image features to predict scene graphs. For HOI detection, we deploy a transformation network to transfer knowledge in scene graphs into HOI representations and then leverage another decoder network to predict HOIs. Additionally, we further devise a knowledge-transfer component for HOI queries via a cross-attention technique. 
}


In summary, our contributions are {four}-fold:

\begin{itemize}

    \item  {For the first time,  we propose a unified one-step model that simultaneously addresses scene graph generation and human-object interaction detection, both of which are cohesively trained in an end-to-end manner.}

	\item {We propose learning a holistic semantic-spatial representation with bounding-box-based image segmentation to explicitly incorporate both semantic and spatial information for the prediction of SGs and HOIs.}
	
	\item {We devise feature and query transformation modules to transfer knowledge present in SGs to HOIs, thereby bridging the gap between these two tasks.}
	
	\item We evaluate our method on a suite of benchmark SGG and HOI datasets: Visual Genome, V-COCO, and HICO- DET. On a wide range of evaluation tasks, our proposed SG2HOI+ method outperforms the state-of-the-art one-stage SGG models and achieves performance competitive with state-of-the-art HOI methods.
\end{itemize}

It is important to highlight that this paper is a substantial extension of our preliminary work accepted at ICCV 2021~\cite{he2021exploiting}. The earlier version primarily centered on HOI detection by leveraging existing scene graph models. In this current iteration, we have undertaken substantial enhancements and extensions as below:

\begin{itemize}
	\item We have developed a distinctly novel one-step model founded upon the Transformer architecture, contrasting with our prior two-step method.
 
	\item We jointly learn scene graph generation and human-object interaction detection, instead of simply utilising scene graphs from off-the-shelf SGG models.
	
	\item To enable the transferring from  SGs to HOIs, we carefully develop feature and query transformation modules, rather than the message-passing technique in our previous work. 
	
	\item {We conduct more experiments on the both SGG and HOI datasets. }  
	Our newly proposed method is considerably superior to our previous SG2HOI in both model performance and time efficiency. 
 \end{itemize}
 

\section{Related Work}\label{sec:related}

\noindent {\textbf{Scene Graph Generation (SGG)}, also referred to as visual relationship detection, is a fundamental task in computer vision that aims to detect and localize relationships between objects present in an image. The early approaches to SGG, such as those introduced by Lu et al. \cite{lu2016visual}, Zhang et al. \cite{zhang2017visual}, Xu et al. \cite{xu2017scene}, Newell et al. \cite{newell2017pixels}, and Dai et al. \cite{dai2017detecting}, followed a conventional two-stage pipeline. This paradigm used an off-the-shelf object detection network to generate object proposals and subsequently employed a relation prediction module to classify the combined subject-object pairs. For instance, Lu et al. \cite{lu2016visual} introduced the use of language priors to refine visual features of detected objects. Zhang et al. \cite{zhang2017visual} modeled relation predictions as vector translations through a visual translation embedding network. Xu et al. \cite{xu2017scene} proposed an iterative message-passing technique using standard RNNs to refine relation predictions. Later on, an issue of extremely biased predicate distribution was revealed in the Visual Genome dataset \cite{zellers2018neural}. } This discovery prompted a series of subsequent works to address the bias problem in SGG. For instance, Tang et al. \cite{tang2020unbiased} analyzed the cause-effect relationship during inference and devised a novel debiasing strategy that could be seamlessly applied in other SGG models. He et al. \cite{he2020learning} tackled the problem by introducing feature hallucination and knowledge transfer from head relations to tail relations. However, a common thread among these works is the independent learning of object visual features and relation features.

Recently, inspired by the remarkable success of the transformer architecture in natural language processing \cite{vaswani2017attention}, researchers have extended it to computer vision tasks, such as DETR \cite{carion2020end} and ViT \cite{dosovitskiy2020image}. The core concept behind these approaches is to frame task-specific classification problems as set prediction challenges. This idea offers the advantage of modeling the entire model in a one-stage fashion, significantly enhancing inference speed. In this context, Liu et al. \cite{liu2021fully} pioneered the use of a fully convolutional network to generate scene graphs within a one-stage model. Subsequently, Li et al. \cite{li2021sgtr} introduced transformers to scene graph generation by constructing bipartite graphs. Cong et al. \cite{cong2022reltr} embraced the perspective of scene graph generation as a set prediction task and formulated the end-to-end RelTR model, utilizing an encoder-decoder architecture.


 \noindent \textbf{Human-Object Interaction (HOI)} is concerned with the identification and localization of interactions of human-object pairs, requiring an in-depth understanding of the scene. The majority of prior works \cite{qi2018learning,gupta2019no,ulutan2020vsgnet,gao2018ican,liao2020ppdm,gkioxari2018detecting,zhang2021mining} adhere to a two-step paradigm. The initial step entails employing a pre-trained object detection network, such as Faster-RCNN \cite{ren2016faster}, to generate human and object proposals, thereby yielding a quadratic number of human-object pairs. Subsequently, these pairs are channeled into an interaction classification module. However, many studies focus on the second stage, aiming to delve into richer visual and contextual cues for HOI recognition.  
For instance, VSGNet \cite{ulutan2020vsgnet} centered on relative spatial and structural cues, developing two modules: a spatial attention network and an interactiveness graph. TIK \cite{li2019transferable} introduced an interactiveness network to suppress non-interactive human-object pairs, thereby enhancing the performance of HOI models.  PD-Net \cite{zhong2020polysemy} devised a Polysemy Deciphering Network to handle the diverse semantic interpretations of verbs. CHGNet \cite{wang2020contextual} developed a homogeneous graph network, facilitating message passing between both homogeneous and heterogeneous entities. However, the propagation of messages is conducted in an agnostic manner. To address the issue of insufficient and indistinguishable visual appearance features for different interactions, FCNNet \cite{liu2020amplifying} proposed a multi-stream pipeline. This approach seeks to amplify critical cues, including object label embeddings (word2vec), fine-grained spatial layout information, and flow predictions, ultimately enhancing the discernment of various interaction types.

Nonetheless, these methods often entail processing an extensive number of human-object pairs, a large proportion of which are non-interactive. This can severely impact computational efficiency. To solve this issue, several efforts \cite{kim2020uniondet, wang2020learning,zou2021end, chen2021reformulating, kim2021hotr} have been directed towards the development of end-to-end HOI detection paradigm. For instance, InterPoint \cite{wang2020learning} introduced a fully convolutional approach that concurrently detects interaction points and predicts interactions. This approach bypasses the need to compute all possible human-object pairs, thereby enhancing computational efficiency.
Subsequent works \cite{kim2021hotr,zou2021end,chen2021reformulating}, drawing inspiration from the principles of DETR \cite{carion2020end}, propose to formulate HOI prediction as a set prediction task. Although SGG and HOI exhibit distinct focal points – SGG emphasizing relationships among various object pairs and HOI targeting human-centering interactions – we assert that scene graphs can serve as pivotal cues for HOI detection when both tasks are jointly modeled in an end-to-end network.


\noindent \textbf{Multi-task Learning (MTL)}, a concept introduced by Caruana et al. \cite{caruana1997multitask}, pertains to the simultaneous learning of multiple interconnected tasks within a single network, with joint training of distinct tasks. The principal advantage of MTL lies in its potential to facilitate knowledge sharing among tasks, thereby enhancing the model's generalizability. Recent times have witnessed the wide application of MTL across a diverse spectrum of domains, such as natural language processing \cite{clark2019bam,liu2019improving} and computer vision \cite{he2017mask,ghiasi2021multi,xiao2018unified,kokkinos2017ubernet}.
Prominently, the widely acclaimed MTL network Mask-RCNN \cite{he2017mask} combined object detection, instance segmentation, and keypoint detection into a unified framework, eclipsing models exclusively trained on each individual task. The success of Mask-RCNN is attributed to the underlying shared characteristics of these tasks, such as object localization information, which enable their effective joint learning. Concurrently, Kokkinos \cite{kokkinos2017ubernet} proposed a unified convolutional network for training low-, mid-, and high-level vision tasks in an end-to-end manner, yielding state-of-the-art performance across several vision tasks. More recently, Ghiasi et al. \cite{ghiasi2021multi} introduced a multi-task self-training (MuST) approach that leverages knowledge learned from teacher models to enhance the training of a singular, versatile student model.

{
 In this work, we hold the perspective that SGG and HOI detection inherently possess shared knowledge, not only in terms of their outputs but also in their underlying representations. Thus, our aim is to devise a unified network with joint training for both tasks, each of which can mutually benefit from each other. 
 }
\section{Method}
\begin{figure*}
	\centering
	\includegraphics[width=1\linewidth]{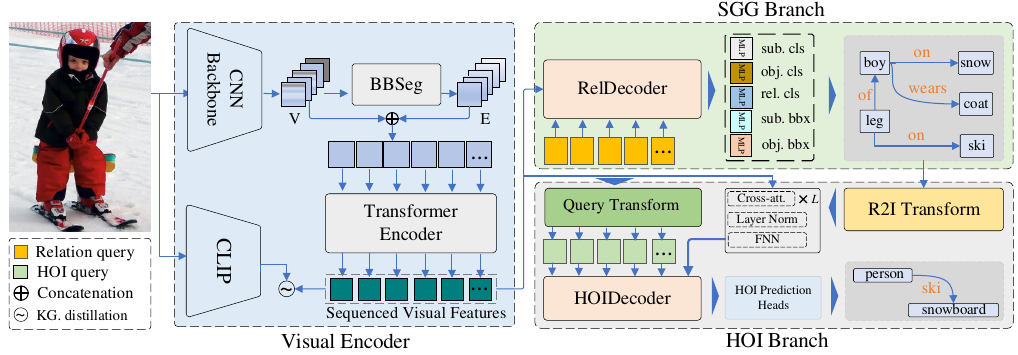}
	\caption{ {
 The overview of our proposed unified end-to-end framework for SGG and HOI detection comprises three key modules: bounding box-based segmentation, relation transformer, and interaction transformer. The bounding box-based segmentation network generates holistic semantic-spatial features to enhance relationship and interaction predictions. The relation transformer generates scene graphs from a set of image features. Last,  we employ two transformer-based networks $\mathrm{R2ITrans}$ and $\mathrm{HOIDecoder}$  to convert relationships to interactions and predict HOIs, respectively. }
 }
	\label{fig:framework}
\end{figure*}

{
In this work, we attempt to
 integrate the two  tasks of SGG and HOI detection within a single network. Given an input image, our model is designed to simultaneously predict SGs and HOIs. 
 To achieve this, we formulate the two tasks as sequential set prediction problems, employing two transformers-based networks.
Our overall framework, illustrated in Figure \ref{fig:framework}, encompasses three key modules: bounding box-based segmentation, relation transformer, and interaction transformer. Specifically, the bounding box-based segmentation network aims to generate holistic semantic-spatial features to enhance the prediction of relationships and interactions. The relation transformer takes as input the encoded image features to generate scene graphs. Finally, for HOI detection, we deploy   two additional sub-networks  $\mathrm{R2ITrans}$ and $\mathrm{HOIDecoder}$. Specifically, $\mathrm{R2ITrans}$ is a transformer-based encoder network, which aims to transform the predicted scene graphs into HOI representations and then a $\mathrm{HOIDecoder}$ is deployed to predict HOIs.
}

\subsection{Holistic semantic-spatial feature learning}
As broadly evidenced by many HOI and SGG arts~\cite{zellers2018neural,tang2019learning,he2021exploiting,iftekhar2022look,liu2020amplifying}, it is widely recognized that semantic and spatial information plays a crucial role in enhancing the prediction of relationships and interactions. In two-stage literature~\cite{zellers2018neural,tang2020unbiased, he2020learning}, a popular way is to generate a spatial map via a pair of binary maps using the bounding boxes from an object detection network, while the semantics mainly stems from word embeddings (e.g., Glove \cite{pennington2014glove}) of object labels.  

However, in the context of one-stage approaches, these techniques become infeasible due to the inability to access label and bounding box information in advance. Drawing inspiration from image segmentation methodologies \cite{minaee2021image}, which naturally yield segmentation maps with both rich semantics and locative information, we introduce the concept of learning a holistic semantic-spatial feature map through bounding box-based segmentation.


\noindent \textbf{Bounding box-based segmentation (BBSeg):} Different from the conventional image segmentation that learns accurate pixel-level segmentation, our BBSeg tries to produce segmentation at bounding box-level. There are two main reasons: (1) the segmentation annotations in the SGG or HOI datasets  (e.g., VG \cite{krishna2017visual} or HICO-DET \cite{chao2018learning}) are unavailable and manually labeling them is both time- and labor-consuming; (2) for the relation and interaction prediction, the pixel-level spatial cues may have a trivial contribution because both tasks focus more on the contextual or relative spatial information.  

 Given an image $\boldsymbol{x}_{i} \in \mathcal{D}$,  we first use a CNN backbone to extract the visual features $\boldsymbol{V}_i\in \mathbb{R}^{H\times W \times C}$, where $H$, $W$ and $C$ denote the height, width and channel size of a feature map, respectively. Then, following previous works~\cite{carion2020end,zou2021end}, we  deploy  $5$ convolutional layers with the kernel size $1 \times 1$   to reduce the channel dimension of $\boldsymbol{V}_i$ to $C^\prime$, i.e., $\boldsymbol{V}_i\in \mathbb{R}^{H\times W\times C^\prime}$.

 Following the mainstream image segmentation~\cite{minaee2021image}, we manipulate a fully convolutional network $\mathrm{FCN_{seg}}$ to predict the bounding box-based segmentation $\textbf{s}_i \in  \mathbb{R}^{H \times W \times {(N^o+1)}}$ as following: 
 \begin{equation}
  \textbf{s}_i=\mathrm{ FCN_{seg}}(\boldsymbol{V}_i)
 \end{equation}
where $N^o+1$ means the number of object categories in the dataset and an extra background class.   During training, we treat the annotated bounding box of an object as the ground truth of bounding box-based segmentation. That is, we label all pixels in a bounding box as the corresponding object class, and thus we can obtain a ground truth segmentation label $\textbf{y}_i \in \{0,1\}$ for image $i$, i.e., $\textbf{y}_i \in \mathbb{R}^{H^\prime \times W^\prime \times {(N^o+1)}}$ where $H^\prime$ and $W^\prime$  are the height and weight of the image $i$. Then, we use the binary cross-entropy loss to learn $\textbf{s}_i$, as  below:
\begin{equation}
 \mathcal{L}_{seg} = \sum_{i} \sum_{j=1}^{N^{o}+1} (\mathrm{BCE}(\textbf{s}_i[:,:,j],\textbf{y}_i[:,:,j]))
\end{equation}
where $\textbf{s}_i[:,:,j]$ means that we slice $\textbf{s}_i$ in the channel dimension. 
It is worth noting that during calculating  $\mathcal{L}_{seg}$, we  rescale $\textbf{y}_i$ to the same size with $\textbf{s}_i^{(j)}$ by downsampling.

\noindent \textbf{Semantics embedding:} If $\textbf{s}_i$ solely consists of the discrete number zero and one,  it lacks essential semantic information. To mitigate this problem, we enhance each position whin $\textbf{s}_i$ with text embeddings sourced from CLIP~\cite{radford2021learning}. More concretely, based on the prediction score in each position $(h,w)$ where $1\leq h \leq H$ and $1\leq w \leq  W$, we can obtain a semantic map that indicates the object category of each position.    Then, we use the text encoder in CLIP to embed a prompt ``A photo of a \texttt{[object class]}" into a semantic vector $\boldsymbol{e}_{(h,w)} \in \mathbb{R}^{C^\prime}$. Notably, if the object category corresponds to the background, we adopt a $\boldsymbol{0}$ vector as the embedding. By doing this, we acquire $(H\times W)$ text embeddings which are then reshaped into the holistic semantic-spatial map denoted as $\boldsymbol{E}_s \in \mathbb{R}^{H\times W \times C^\prime}$.

\noindent \textbf{Feature aggregation:} Now we  have the visual feature $ \boldsymbol{V}_i $ and the holistic semantic-spatial $\boldsymbol{E}_s$, and we aggregate them together into a comprehensive representation:
\begin{equation}
	\boldsymbol{f}^{agg} = \mathrm{CONCAT}( \boldsymbol{W}^v \boldsymbol{V}, \boldsymbol{W}^s \boldsymbol{E}) 
\end{equation}
where $\boldsymbol{W}^v$ and $\boldsymbol{W}^s$ are two transformation layers implemented by a convolutional layer  with $1\times1$ kernel size, and $\mathrm{CONCAT}$ means the concatenate operation in the channel dimension, i.e.,  	$\boldsymbol{f}^{agg} \in \mathbb{R}^{H\times W \times d}$ where $d=2\times C^\prime$. 

\subsection{Scene graph generation} \label{sec:vsgg}
Our relation prediction is built on a standard transformer architecture with an encoder and decoder network, both of which consist of a couple of multi-head self-attention modules and feed-forward networks(FFN). However, different from the previous one-stage paradigms~\cite{liao2022gen,kim2021hotr}, we deploy an additional visual-linguistic(VL) distillation head on the encoded features, aiming to impart the encoded features with rich knowledge from the large-scale pretrained   VL models, e.g., CLIP~\cite{radford2021learning}. We believe this distillation would guide the encoder network to learn better representations, and to some extent,  alleviate the biased relation prediction in SGG because of more injection of generic VL knowledge from CLIP.   

\noindent \textbf{Encoder:} Our encoder,  based on a transformer network, takes as input a set of feature vectors $\boldsymbol{f}^{agg}$. {Concretely, we first flatten the feature map $\boldsymbol{f}^{agg}$ into a sequence of vectors with the length of $(H\times W)$.  Additionally, we add a  start token $\texttt{[CLS]}$  at the first position of the flattened vectors to present the global feature of the image. } In our implementation, the input feature of $\texttt{[CLS]}$ is the global image feature produced by a global average pooling layer (GAP)~\cite{ulutan2020vsgnet}.  Then, we add a position encoding $\boldsymbol{p} \in \mathbb{R}^{(H\times W+1) \times d}$ to encode the position of each vector in the sequence. 
Mathematically, we could formulate the encoding process as:
\begin{equation}
	 \boldsymbol{f}^{en} \!=\! \mathrm{Encoder}([\mathrm{GAP(\boldsymbol{V})},\boldsymbol{f}^{agg}_1,\boldsymbol{f}^{agg}_2, \cdots, \boldsymbol{f}^{agg}_{H\times W}],\boldsymbol{p})
	 \label{eq:reltrans}
\end{equation}  
where $[;]$ means the concatenation operation.

Since the visual relationships distribute unevenly in datasets, e.g., VG~\cite{krishna2017visual}, the model can readily get stuck in the dilemma of learning visual patterns for the data-starving categories,  which, in turn, can degrade the performance of downstream tasks like SGG and HOI detection.  { To remedy this, we draw inspiration from the achievements of pretrained VL models~\cite{radford2021learning} and propose the utilization of a large-scale pretrained VL model to guide our feature encoding process. This operation aims to align the encoded features with the comprehensive knowledge captured by the pretrained visual-linguistic model, ultimately enhancing the encoded representations.
In practice, we implement this guidance mechanism based on two types of features: (1) the output of the \texttt{[CLS]} token; and (2) the pooled feature of  Eq.(\ref{eq:reltrans}) as \cite{liao2022gen}. 
Specifically,  we embed an image into a vector $\boldsymbol{f}^{clip}$ using the visual encoder of CLIP, and employ an $\mathcal{L}_{algn}$ loss to minimize the discrepancy between $\boldsymbol{f}^{clip}$ and $\boldsymbol{f}^{en}{\texttt{[CLS/pool]}}$:} 
\begin{equation}
	{\mathcal{L}_{algn} = \left| \boldsymbol{f}_{clip} - \boldsymbol{f}^{en}_{\texttt{[CLS/pool]}}) \right| }
	 \label{eq:vl_align}
\end{equation}
{where $\boldsymbol{f}^{en}_{\texttt{[CLS/pool]}}$ denote the two types of features: the output feature of the token $\texttt{[CLS]}$ in Eq.(\ref{eq:reltrans}) or the pooled feature  of Eq.(\ref{eq:reltrans}), i.e., $\boldsymbol{f}^{en}_{\texttt{[pool]}} = \frac{1}{H\times W}\sum f_i^{agg} $.}

\noindent \textbf{Relation prediction:} We  use a query-based technique~\cite{tamura2021qpic,carion2020end} to predict  SGs. Specifically, our decoder encompasses several multi-head cross attention modules and takes as input the encoded feature  $\boldsymbol{f}^{en}$ and a set of learnable queries vectors $\boldsymbol{Q}^{rel} \in \mathbb{R}^{N^{rel}\times d}$ where $N^{rel}$ is the number of relation queries. Note that the position embedding $\boldsymbol{p}$ is also added in $\boldsymbol{Q}^{rel}$. The decoding process can be written as:
\begin{equation}
	\boldsymbol{f}^{sg} = \mathrm{ReLDecoder}(\boldsymbol{f}^{en}, \boldsymbol{Q}^{rel})
	\label{eq:rel_de}
\end{equation}

For the output scheme, we represent each edge in an SG as a $5$-tuple denoted as $(b^s, l^s, r, b^o, l^o)$, indicating the subject box, subject label, relationship, object box, and object label, respectively, following  \cite{tamura2021qpic,chen2021reformulating}. To predict these components, we employ five parallel branches implemented using Multi-Layer Perceptrons (MLPs).   Note that the predicted bounding box is a 4-D box presented by a center point $(x,y)$, height $h$, and width $w$.  
 The relation branch generates a $(N^r+1)$-dimensional score distribution for relations, where $N^r$ denotes the number of relation predicates, along with an additional class representing non-relations.

In practice, we employ two strategies for initializing the object and relation classifiers. the conventional way is to randomly initialize the classifier weights, whilst the other way is to use the text embedding from CLIP~\cite{radford2021learning} to initialize the weights. The latter has demonstrated promising improvements across a diverse spectrum of tasks \cite{liao2022gen,li2022clip,Wang_2022_CVPR}.
For the training loss, we elaborate on it in the $\S\ $\ref{sec:loss}.
 

\subsection{Predicting HOIs from SGs}
 Given a set of predicted relation tuples, our goal is to predict HOIs from them.  {Our motivation is that an SG can provide the HOI prediction explicitly or implicitly with essential high-level knowledge 
 \cite{he2021exploiting}.  For instance, the relationship ``sitting on" can directly correspond to the interaction of the same name. However, some interactions might require inference from the combination of multiple relationships. For instance, the interaction ``working on computer" can be acquired from the relation triples of $<$man in front of computer$>$ and $<$man sitting on chair$>$. we employ a transformer-based component to translate these relation tuples into HOI-related representations. Subsequently, an interaction decoder is performed to predict HOIs based on these features.}
 
 \noindent \textbf{Relation to interaction transformation:} This component is implemented by a transformer-based encoder $\mathrm{R2ITrans}$ consisting of multiple self-attention layers and feed-forward networks. Precisely, it takes as input the set of relation tuples $\boldsymbol{f}^{sg}$ in Eq.(\ref{eq:rel_de})  and a position encoding $\boldsymbol{p}$.    In formal terms, the transformation processing is written as:
 \begin{equation}
       \boldsymbol{f}^{r2i} = \mathrm{R2ITrans}(  [ \Tilde{\boldsymbol{f}}^{sg}_1, \Tilde{\boldsymbol{f}}^{sg}_2, \cdots \Tilde{\boldsymbol{f}}^{sg}_{N^{rel}} ], \boldsymbol{p})
 \end{equation}
 where $\Tilde{\boldsymbol{f}}^{sg}_k  = \boldsymbol{W}^{t} [\boldsymbol{f}^{sg}_{s_{k}}, \boldsymbol{f}^{sg}_{r_{k}}, \boldsymbol{f}^{sg}_{o_{k}}]$ is the concatenation of the three output hidden features from the subject, relation predicate and object branches, respectively; $\boldsymbol{W}^{t}$ is a transformation layer. 
 
 \noindent \textbf{Feature transformation:} \label{sec:feat_trans} 
{ An intuitive idea is to directly employ an interaction decoder after relation prediction to predict HOIs from $\boldsymbol{f}^{r2i}$, practical results indicate that this approach does not yield promising performance. This is possibly because the HOI prediction solely relies on the relation prediction whose error could completely propagate to the later HOI prediction,  and negatively impact  HOI predictions, particularly during the initial stages of training.}
 To solve this issue, we feed the previously encoded feature  $\boldsymbol{f}^{en}$ in Eq. (\ref{eq:reltrans}) and $\boldsymbol{f}^{r2i}$ together into the HOI decoder.
 In our implementation, we use cross-attention technique to fuse them. Specifically, we treat $\boldsymbol{f}^{en}$   as the queries and $\boldsymbol{f}^{r2i}$ as the keys and values, and let $\boldsymbol{f}^{en}$ cross-attend with $\boldsymbol{f}^{r2i}$. Consequently, we could obtain a set of transformed features $\boldsymbol{F}^{hoi}=\{\boldsymbol{f}^{hoi}_i \in \mathbb{R}^d\}_{i=1}^{H\times W}$.

  \noindent \textbf{Query transformation:}\label{sec:query_trans} {When it comes to HOI queries, we do not independently train a fresh set of randomly initialized queries.
  Instead, we leverage the insights from the earlier established relation queries. We speculate that these relation queries might contain implicit cues related to relationships that can be transferred to interaction queries. This strategy is designed to maximize the alignment between the two tasks.}
  
 To this end,  we again employ the cross-attention mechanism to inject relationship query information into the interaction queries. In practical terms, we start by randomly initializing a set of HOI query vectors denoted as $\boldsymbol{Q}^{hoi}={ q^{hoi}i}{i=1}^{N^{hoi}}$. These HOI query vectors are used to cross-attend the relation queries $\boldsymbol{Q}^{rel}$, which serve as both keys and values. Finally, the transformed $\boldsymbol{Q}^{hoi}$, along with the position encoding $\boldsymbol{p}$, is fed into the HOI decoder, and the process is represented as:
\begin{equation}
\boldsymbol{h}_{hoi} = \mathrm{HOIDecoder}(\boldsymbol{F}^{hoi},\boldsymbol{Q}^{hoi})
\end{equation}

   \noindent \textbf{HOI prediction:} 
   In our implementation, an HOI is represented by a 5-tuple consisting of the following components: human box, human score, interaction, object box, and object label. This representation is implemented through five distinct MLP branches. The noteworthy distinction from the previous relation classification lies in the alteration of the subject label classification, which is changed into a binary human classification.
  
 \subsection{Training and Inference} \label{sec:loss}
 
 In our model, we incorporate four objective functions: $\mathcal{L}{seg}$, $\mathcal{L}{dis}$, $\mathcal{L}{rel}$, and $\mathcal{L}{hoi}$, which correspond to the holistic semantic-spatial map, visual-linguistic knowledge distillation, relation prediction, and HOI prediction, respectively. While the details of the former two have been previously discussed, we will now delve into further elaboration on the latter pair.

In accordance with prior one-stage methodologies~\cite{zou2021end,chen2021reformulating}, the computation of $\mathcal{L}_{rel}$ and $\mathcal{L}_{hoi}$ involves two primary steps: bipartite matching to select paired instances and subsequent loss calculation for these matched instances. Similar to \cite{carion2020end}, we utilize the Hungarian algorithm \cite{kuhn1955hungarian} for determining the matching cost between predictions and ground truth instances. Thereafter, the matched instances are treated as ground truth, forming the basis for loss calculation during backpropagation. For instance, regarding the relation matching cost, our objective is to utilize the minimum cost for assigning $N^{gt}$ ground truth relation $5$-tuples to the $N^{rel}$ output instances, where $N^{gt} << N^{rel}$. It is important to note that we do not compute the loss for unmatched pairs, i.e., the remaining $N^{rel} - N^{gt}$ pairs. Each matching cost for a $5$-tuple encompasses five components: subject and object box matching loss, subject and object label classification loss, and relation label classification loss. Mathematically, the relation match cost  can be formulated as:
 \begin{equation}
 	 \mathcal{L}_{rel}^{cost} = \alpha_1 L_{box}^\star  + \alpha_2 L_{cls}^\star  + \alpha_3 L_{cls}^r
 	 \label{eq:loss_cost}
 \end{equation}
where $\star$ represents either the subject or object entity and $\alpha_{1\sim3}$ are hyperparameters employed to modulate the influence of the three corresponding terms.  Followed \cite{zou2021end}, the box matching loss encompasses a box regression loss $L_1$ and an intersection-over-union loss $L_{iou}$, while the classification loss is calculated by a standard cross-entropy loss. After identifying the best-matched pairs,  we treat their matching cost as $ \mathcal{L}_{rel}$.

Analogously, we adopt the same strategy to compute $\mathcal{L}_{hoi}$. The overall loss is formulated as follows:
\begin{equation}
	\mathcal{L} = \lambda_1 \mathcal{L}_{seg} + \lambda_2 \mathcal{L}_{algn} + \mathcal{L}_{rel} + \mathcal{L}_{hoi}
	\label{eq:loss_all}
\end{equation}
where $\lambda_1$ and $\lambda_2$ function as hyperparameters for regulating the relative contributions of the respective terms.

It is worth noting that since our training datasets do not contain both     relation and interaction annotations in an image, at the training stage, we alternatively train the two tasks. For more training details, please refer to the section of implementation detail $\S\ $\ref{sec:imp_de}.

  

\section{Experiments}
In this section, we comprehensively evaluate our model, denoted as \textbf{\textbf{SG2HOI+}}, across two distinct tasks: SGG and HOI detection, across several benchmark datasets. For SGG, following \cite{zellers2018neural,he2021exploiting}, we test it  on the mainstream visual relation detection dataset Visual Genome (VG)~\cite{krishna2017visual}, while performing  HOI  on the two public benchmark datasets: V-COCO \cite{gupta2015visual} and HICO-DET~\cite{chao2018learning}. In the following subsections, we carefully elaborate on our datasets, evaluation metrics, main results on the two tasks, and the ablation study.  Last, we visualize some cases of   predicted SGs and HOIs by our method.

\subsection{Datasets}
\subsubsection{Visual Genome (VG) \cite{krishna2017visual}}  is a mainstream benchmark  dataset  for scene graph generation. Following the widely adopted split~\cite{zellers2018neural,tang2020unbiased}, 
the training, validation and testing set consist of $57,723$, $5,000$ and $26,443$ images, respectively. Additionally, we choose the   most frequent $150$ object classes and $50$ relationship predicates  to generate scene graphs. 

\subsubsection{V-COCO\cite{gupta2015visual}}  is split into training, validation and test sets with $2,533$, $2,867$ and $4,946$ images, respectively.
Following previous works \cite{ulutan2020vsgnet,wang2020learning}, we also use the training and validation set, containing $5,400$ images in total, to
train our model. Each human-object pair in V-COCO is labeled with a $29$-D one-hot vector. It is worth noting that
among the $29$ actions, three (cut, hit, eat) have no interaction objects.  

\subsubsection{HICO-DET\cite{chao2018learning}}
  is annotated with $600$ human-object interaction classes, $80$ object classes and $117$ actions, including
  a no-interaction class. Following previous works \cite{ulutan2020vsgnet,wang2020learning,chao2018learning},
  we categorize the interactions into three groups: Full, Rare and Non-Rare, based on the number of its training samples.
  Also, we further conduct the experiments under a ``Known   Objects'' setting but the default is unknown for objects.

\subsection{Baselines}
{
To thoroughly demonstrate the efficacy of our proposed model, we have selected a set of representative state-of-the-art models for SGG and HOI detection as our comparative baselines. To ensure a fair and comprehensive comparison, we primarily focus on assessing against one-stage SGG and HOI models, while also presenting results in comparison with competitive two-stage approaches. 
}

{
As for SGG models, we divide them into two groups: two- and one-stage. The conventional two-stage methods include  Seq2seq~\cite{lu2021context}, Motifs~\cite{zellers2018neural}, TDE~\cite{tang2020unbiased}, VGTree~\cite{tang2019learning}, IMP~\cite{xu2017scene}, while the one-stage models have SGTR~\cite{li2021sgtr}, FCSGG~\cite{liu2021fully}, and RelTr~\cite{cong2022reltr}.  It is noteworthy that Seq2seq~\cite{lu2021context} is classified as a two-stage approach due to its utilization of a pre-existing object detection network.
}

 {
In terms of HOI models, we also divide them into two categories: two- and one-stage. The former models include InteractNet~\cite{gkioxari2018detecting}, FCMNet~\cite{liu2020amplifying},  TIN~\cite{li2020pastanet}, CHG~\cite{wang2020contextual}, GPNN~\cite{qi2018learning}, iCAN~\cite{gao2018ican}, VSGNet~\cite{ulutan2020vsgnet}, No-Fills~\cite{gupta2019no}, IDN~\cite{li2020hoi} SCG~\cite{zhang2021spatially}, ATL~\cite{hou2021affordance}, and SG2HOI~\cite{he2021exploiting}. However, for the one-stage HOI models, we further  categorise  them into two subgroups: two- and single-decoder based methods,
according to whether or not to use two decoders for object and interaction prediction. For example, the two-decoder based one-stage HOI models have UnionDet~\cite{kim2020uniondet}, HOTR~\cite{kim2021hotr}, AS-Net~\cite{chen2021reformulating}, GEN-VLKT \cite{liao2022gen}, while  the single-decoder models contain IP-Net~\cite{wang2020learning},  QPIC~\cite{tamura2021qpic}, SSRT \cite{iftekhar2022look}, and HOI-Trans~\cite{zou2021end}.
}
 
\subsection{Evaluation Metrics}

\noindent\textbf{SGG:} There are three  mainstream subtasks in the SGG evaluation: (a) Predicate Classification (PredCls)  with ground-truth object labels and boxes provided; (b) Scene Graph Classification  with only the labels provided; and (c) Scene Graph Detection  with neither of them. However, since our model simultaneously predicts the bounding box, object class, and predicate, at inference, we cannot obtain the bounding box or the corresponding object label in advance. Hence, in this paper, we only report the results of the most challenging task  SGDet, as SGTR\cite{li2021sgtr}. 

We report two types of metrics: the unbiased metric mean Recall@K ($\mathbf{{mR@K}}$)  and the conventional metric $\mathbf{{R@K}}$. It is worth noting that the latter one does not reflect a model's true performance on tail relations~\cite{chen2019knowledge,tang2020unbiased}.  Following \cite{chen2019knowledge}, we also report the results under two scenarios: graph constraint and unconstraint. 

\noindent \textbf{HOI:} We report the results on the standard
evaluation metric \cite{kim2021hotr,ulutan2020vsgnet} role Mean Average Precision
(mAProle). More concretely, if an HOI triple meets the following two conditions: (1) both bounding boxes of the detected human and object are greater than $0.5$ with the respective annotated ground-truth boxes; and (2) their interaction class is correctly predicted, then we consider the HOI triple is correct.

\subsection{Implementation Details} \label{sec:imp_de}
Following \cite{zou2021end}, we initialize our model's parameters using the pretrained DETR \cite{carion2020end} model on the MS-COCO dataset. We set the number of relation queries and interaction queries to $100$, which means the upper limit of predicted  relation triples and HOIs is $100$ for an image.  Following \cite{zou2021end}, we set the three hyperparameters $\alpha_{1\sim3}$ in Eq.(\ref{eq:loss_cost})   to $1$, $2$ and  $4$. The hyper-parameters $\lambda_{1\sim2}$ a in Eq.(\ref{eq:loss_all}) are set to $2$ and $4$. For an input image, we  leverage some basic augmentation techniques, such as $50$\% probability to flip the image and adjust the brightness and contrast, as \cite{zou2021end}. The architecture of the segmentation module consists of $5$ convolution layers with a kernel size of $3 \times 3$ and batch norm operation.    The last layer is a classification layer with a kernel size of $1 \times 1$. The dimension of semantic embeddings from CLIP is set to $512$.   We use the pretrained CLIP encoder of VIT-B/$32$  to embed  visual features. Following CLIP, we rescale the input image into $224 \times 224$. Note that during training,  the parameters of the visual encoder and text encoder of CLIP are frozen. In default, we choose $\boldsymbol{f}^{en}{\texttt{[CLS]}}$ to align with the corresponding CLIP feature.  {
The $\mathrm{encoder}$ encompasses a  $6$-layer transformer-based blocks  and the following $\mathrm{RelDecoder}$ is a relation decoder with   $6$-layer transformer-based blocks.   The MLP module to predict the  bounding box consists of $3$ non-linear project layers.  The query and feature transformation modules   are $2$-layers cross-attention layers.
The $\mathrm{R2ITrans}$ is a $3$-layer transformer and the following   cross-attention component consists of two multi-head attention layers, and  $\mathrm{HOIDecoder}$ is $6$-layer transformer decoder.   
  Since the performance of SGG could influence the downstream task HOI, we initially train the SGG module for $20$ epochs using AdamW optimizer with a weight decay of $10^{-4}$. At this stage, the learning rate of  transformer networks is set at $10^{-4}$, while the backbone's learning rate is set to $10^{-5}$. Subsequently, the HOI module is jointly trained with a learning rate of $10^{-4}$, while the SGG part's learning rate is set at $10^{-5}$. When employing CLIP class embeddings as classifiers, the learning rate is adjusted to $10^{-5}$. All experiments are conducted using a batch size of $8$ on $8$ Tesla V$100$ GPUs with CUDA $11.6$.
  }

 \subsection{Results on SGG}
  {We first evaluate our model on the task of SGG  and categorize the comparison methods into two groups: conventional two-stage and one-stage approaches. The two-stage methods typically employ a pre-existing object detection network to pre-extract object proposals and then  classify the subject-object pairs into relationships. On the contrary, the one-stage models simultaneously optimize object detection and relation classification as part of a single process. Additionally, we present results for two additional scenarios: the long-tail split and zero-shot SGG, as  \cite{li2021sgtr}.}
 
 
  \noindent \textbf{Main results of SGG:}  Table \ref{tab:sgg} presents a comprehensive comparison of results across various evaluation metrics, including Recall@K and mRecall@K, where $K \in \{20, 50, 100\}$. We provide results for four distinct variants of our model: $\textbf{SG2HOI+}_{rand}$, where the object and predicate classifiers are initialized randomly; $\textbf{SG2HOI+}_{clip}$, utilizing embeddings from CLIP for initialization of these classifiers; $\textbf{SG2HOI+}_{cls}$, aligning the output feature $\mathrm{[CLS]}$ with the CLIP feature for VL feature alignment; and $\textbf{SG2HOI+}_{pool}$, employing the pooled feature from Eq.(\ref{eq:reltrans}) to align with the CLIP feature. Importantly, the latter three variants employ embeddings from CLIP to initialize their object and predicate classifiers. A detailed analysis of the results reveals the following observations:
 
  (1) The two-stage approaches consistently exhibit a substantial performance advantage over the majority of one-stage methods in both Recall@K and mRecall@K metrics, particularly with regards to Recall@K. For instance, MOTIFS demonstrates an approximate $26$\% improvement over FCSGG on average across the constrained and unconstrained scenarios. This trend may arise from the fact that two-stage methods can easily leverage external knowledge sources such as statistical frequency priors or language priors, as well as exploit localization cues, to enhance relation classification.
However, due to the absence of access to object detection results in advance, one-stage methods struggle to effectively incorporate semantic and spatial cues for relation recognition. For instance, methods like FCSGG can only rely on visual features. In contrast, our model exploits these cues through a holistic semantic-spatial feature produced by a bounding box based segmentation  module. As a result, the holistic feature plays a role in narrowing the performance gap between the two-stage and one-stage methods. As observed, our best model  ($\textbf{SG2HOI+}_{pool}$) surpasses the majority of one-stage models, for example, exhibiting an average improvement of $1.3$ points over SGTR.
 
 
 (2) The classifiers initialized with enriched pre-trained visual-language knowledge demonstrate a notable advantage over the conventional random initialization approach. For instance, in a constrained scenario, the CLIP-based model (SG2HOI+$_{clip}$) outperforms the randomly initialized model (SG2HOI+$_{rand}$) by an average margin of $2$ points for Recall@K and $3.5$ points for mRecall@K. This superiority can be attributed to the fact that the incorporation of visual-language (VL) knowledge enhances the model's ability to generalize, and to some extent, alleviates the bias inherent in SGG. Consequently, the performance improvement is more pronounced in the unbiased mRecall@K metric.
Moreover, as we incorporate the visual-language knowledge distillation module within the feature encoding stage (see Section \ref{sec:vsgg}), the resulting learned features become more aligned with CLIP  embeddings. Consequently, utilizing CLIP embeddings as the classifier could readily   classify them, leading to improved performance.


\noindent (3) Although   most of the one-stage methods do not achieve SOTA performance, they show superior inference time. Note that during the test time efficiency of SGG, we do not calculate the time cost of HOI prediction. All comparison methods are run in the same environment with one V100 GPU and the batch size is set to 1.  Specifically, from the results of time cost in the rightmost column of  Table \ref{tab:sgg}, we could find that the average inference time of one-stage method is $4\sim5$ times faster than the two-stage method when all of them use the same backbone ResNet-101.   The main reason is that those one-stage methods directly predict relationship triple without the time-consuming post-processing, such as NMS~\cite{he2016deep} or enumerating subject-object pairs.  

\noindent \textbf{Zero-shot  results:}  To  text our model performance on the zero-shot scenario introduced by \cite{tang2020unbiased}, we select  unseen relation triples in the test set as the evaluation samples. Table \ref{tab:sgg} further shows the results of zero-shot SGG in terms of R@$50$/$100$, following the setting in \cite{tang2020unbiased}.  Based on the results, we could see the majority of methods show extremely poor results. For example, many two-stage methods, such as MOTIFS and VCTree, almost gain less than $1$ point performance. In contrast, one-stage methods such as HOTR, which rely exclusively on visual learning, display a higher degree of model robustness and demonstrate better performance. Particularly, our model outperforms the other top-performing model, TDE, by a slight margin of $0.2$ points on average in this zero-shot evaluation.

\noindent \textbf{The long-tail results:}  Following \cite{li2021sgtr}, we divide the predicates into three groups: head (more than $10$k), body ($0.5$k $\sim$ $10$k) and tail (less than $0.5$k)   according to their
instance number in the training set. Table \ref{tab:sgg_frq} show their results on the three groups. Note that  in this setting, we only report the results of mR@K on the graph constraint, the same as \cite{liu2019large}. Note that regarding SGTR, we report our reproduced results based on the released code.   

The results reveal that most of the baseline methods achieve promising performance on the head group but struggle with the tail predicates due to the limited amount of data available. Interestingly, for the tail predicates, one-stage methods demonstrate superior performance compared to two-stage methods. This could be attributed to the fact that one-stage methods do not heavily rely on external knowledge sources such as object label embeddings \cite{zellers2018neural}, thereby exhibiting better generalizability. Notably, even though our model also leverages external knowledge (holistic semantic-spatial features), it derives this knowledge from visual features rather than manual embeddings. This leads to our model demonstrating better generalizability than other two-stage and one-stage methods alike, such as surpassing SGTR by an average of approximately $0.57$ points in terms of mR@$50$.

\begin{table*}[]
	\centering
	\label{tab:sgg}
	\caption{The results  comparison in Visual Genome on the task of \textbf{SGDet}.  We divide the baselines into two types: two-stage and one-stage. $\dagger$ means the results are reproduced by the released code. The best results are bold and the second best result
		is marked with \underline{underline}.}
    \resizebox{2 \columnwidth}{!}{ 
	\begin{tabular}{lll|cc|cc|c|cc}
		\toprule
		&   &    & \multicolumn{2}{c|}{\textbf{Constraint}} & \multicolumn{2}{c|}{\textbf{Unconstraint}}	 & \multicolumn{1}{c|}{\textbf{Zs-SGG}}  & \multirow{3}{*}{\textbf{Time(s)}} \\
		& \textbf{Methods} & \textbf{Backbone} & \textbf{R@20/50/100} & \textbf{mR@20/50/100} & \textbf{R@20/50/100} / & \textbf{mR@20/50/100} & \textbf{R@50/100} \\ \cmidrule{2-9 } 
		\multirow{5}{*}{\textbf{Two-stage}} 
		&  Freq \cite{zellers2018neural}      &    ~~-~~      &     20.1  /  26.2     /  30.1     &  ~4.5     /     ~6.1  /     ~7.1 & ~~-~~ / 25.3 / 30.9 & ~~-~~ / ~5.9 / ~8.9 & 0 / 0 & - \\
		&  IMP \cite{xu2017scene}       &    X-101-FPN       &  14.6 / 20.7 / 24.5  &     ~2.4   / ~3.8 / ~4.8  & 	18.4 /	27.1 /	33.9 &  ~4.3 / ~7.4 / ~9.0  & 0.4 /	0.8& 0.91 \\
		
	&  MOTIFS~\cite{zellers2018neural}    &  X-101-FPN   & 21.4 / 27.2 / 30.3  &  ~4.2 / ~5.7 / ~6.6       & 27.0 /	36.6 /	43.4 & ~~-~~ / ~9.3 / 12.9 & 0.1 / 0.2 &  1.22 \\
		&  VCTree \cite{tang2019learning}      &   X-101-FPN  & 22.0 / 27.9 / 31.3     &     ~5.2 / ~6.9 / ~8.0  & 26.1 /	35.7 /	42.4 &  ~~-~~ / 10.1 /  13.8  & 0.3 /	0.7& 1.35\\
		&   KERN  \cite{chen2019knowledge}    &    VGG16       &     20.6    /    27.1  /   29.8   &   ~4.9   /    ~6.4 /    ~7.3   &  21.8 / 30.9 / 35.8 &   ~6.8 / 11.7 /  16.0  & ~~-~~ / ~~-~~& 1.52\\
	&   TDE \cite{tang2020unbiased}     &  X-101-FPN        &  14.3    / 19.6     /  23.3      &    ~6.3    /   ~8.6  /   10.3   &  14.2 / 23.2 / 27.9 & 12.6 / 15.3 / 19.2 & 2.2 / 2.8& 1.44 \\
	
		&   BGNN\cite{li2021bipartite}      &   X-101-FPN       &    ~~-~~   /    31.0   /   35.8     &   ~~-~~  /    10.7   /     12.6  & ~~-~~ / ~~-~~ / ~~-~~ & ~~-~~ / ~~-~~ / ~~-~~  & ~~-~~ / ~~-~~& 1.29 \\
		&   Seq2Seq$^\dagger$\cite{lu2021context}      &  ResNet-101        &    22.1   /    30.9   /  34.4      &    ~7.5   /  ~9.6    /    12.1 &  26.2 / 30.9 / 37.0  & 11.0 / 14.2 / 18.4  &  0.6 / 1.0 & 1.16\\
		\midrule
		\multirow{8}{*}{\textbf{One-stage}} 
		&     FCSGG~\cite{liu2021fully}    &     HRNet \cite{wang2020deep}     &    16.1   /   21.3   /    25.1   &     ~2.7   / ~3.6   /   ~4.2  & 16.7 / 23.5 / 29.2 &  ~3.8 / ~5.7 / ~7.5 & 0.8 / 1.4& 0.17\\
		&    RelTr~\cite{cong2022reltr}     &     ResNet-50     &        ~~-~~  /   20.2    /  25.2   &    ~--~~  /  ~5.8 / ~8.5 & ~~-~~ / ~~-~~ / ~~-~~& ~~-~~ / ~~-~~ / ~~-~~    & ~~-~~ / ~~-~~  & -\\
	
		&    HOTR$^\dagger$  &     ResNet-101     &        18.2  /    23.5    /   27.7   &    ~7.6   /  ~9.4    /  12.0  & 19.3 / 26.6 / 32.4 &  ~9.0 / ~10.3 /  15.8 &1.0 / 1.6& 0.32 \\
		&    SGTR$^\dagger$ \cite{li2021sgtr}     &  ResNet-101    &     18.5  / 24.6 / 28.4    &    ~8.2  /      \textbf{11.7} / \textbf{14.0} & 22.6 /  28.7 / 33.3 &  10.8 / 11.7 /  17.6  &  1.5 /  2.0 & 0.39\\
		
 
		&  \cellcolor{mygray}\textbf{SG2HOI+}$_{rand}$  &  \cellcolor{mygray}ResNet-101     &  \cellcolor{mygray}17.6 / 23.9 / 27.8    &   \cellcolor{mygray}~5.7 / ~7.6 / 10.1    & \cellcolor{mygray}21.6 / 28.2 / 33.7    &   \cellcolor{mygray}~9.6 / 11.2 / 16.6    & \cellcolor{mygray}2.1 / 2.8  &\cellcolor{mygray}0.28 \\
		&    \cellcolor{mygray}\textbf{SG2HOI+}$_{clip}$  & \cellcolor{mygray}ResNet-50     &      \cellcolor{mygray}18.0 / 24.2 / 28.4   &    \cellcolor{mygray}~6.4 / ~8.5 / 11.8    &   \cellcolor{mygray}22.4 / 28.5 / 33.2    &  \cellcolor{mygray}10.8 / 12.1 / 17.4 &        \cellcolor{mygray}1.9 / 2.7  & \cellcolor{mygray}0.21\\

		&  \cellcolor{mygray}\textbf{SG2HOI+}$_{clip}$  & \cellcolor{mygray}ResNet-101     &    \cellcolor{mygray}{20.5} / {25.3} / {29.5}   &  \cellcolor{mygray}~{8.4} / {11.2} /  {13.7}      &   \cellcolor{mygray}{23.5} / {30.1} / {34.5}    &   \cellcolor{mygray}{11.2 }/ {13.6} / {18.3}      &\cellcolor{mygray}{2.3} / {3.3}  & \cellcolor{mygray}0.28   \\

	 &   {\cellcolor{mygray}\textbf{SG2HOI+}$_{pool}$ } &   {\cellcolor{mygray}ResNet-101}     &      {\cellcolor{mygray}\textbf{21.2}} /   {\textbf{25.9}} /   {\textbf{30.3}}   &    {\cellcolor{mygray}~\textbf{8.9}} /   {\underline{11.4}} /    {\underline{13.9}}      &     {\cellcolor{mygray}\textbf{23.8}} /   {\textbf{30.6}} /   {\textbf{34.6}}    &     {\cellcolor{mygray}\textbf{11.5}} /   {\textbf{13.9}} /   {\textbf{18.6}}      &   {\cellcolor{mygray}\textbf{2.5}} /   {\textbf{3.7}}  & {\cellcolor{mygray}0.28}   \\

		\bottomrule
	\end{tabular} }
 
\end{table*}

\begin{table}[]
	\centering
	\label{tab:sgg_frq}
	\caption{The results of VG on the head, body and tail predicates in terms of mR@K. $\dagger$ means the results are reproduced by the released codes, while $*$ denotes the  results reported in the corresponding papers.  }
		\resizebox{\columnwidth}{!}{
	\begin{tabular}{llccc}
		\toprule
	 \multirow{3}{*}{\textbf{Methods}} & \multirow{3}{*}{\textbf{Backbone}} & \textbf{Head}     & \textbf{Body}     & \textbf{Tail}     \\
		 & & \textbf{ mR@50/100} & \textbf{mR@50/100} & \textbf{mR@50/100} \\ \midrule
	  
		 IMP  & X-101-FPN   &   13.4 / 16.3      &     ~5.6 / ~8.0     &    ~1.3 / 2.0   \\
	
		 Motifs & X-101-FPN &     23.7 / 26.2       &   ~7.5 / ~9.2    &      ~1.5 / 2.1     \\
		 VCTree  & X-101-FPN   &   26.1 / 28.0       &   ~8.0 / ~9.7      &      ~1.6 /  2.0   \\
		 BGNN  & X-101-FPN  &    30.1 / 34.0     &     ~9.7 / 12.9     &    ~1.8 / 2.2    \\ \midrule
		 HOTR$^\dagger$  & ResNet-101  &  23.0 / 25.4      &  13.0 / 16.2       &   3.1 / 4.6     \\
		 SGTR$^\dagger$  & ResNet-101  &     23.6 / 26.2     &   13.7 / 17.8       &   4.0 / \textbf{5.4}      \\
		
		\rowcolor{mygray} \textbf{SG2HOI+}$_{rand}$   & ResNet-101  &       22.4 / 25.3   &  13.1 / 15.6        &      2.9 / 4.2  \\
			\rowcolor{mygray}	  \textbf{SG2HOI+}$_{clip}$   & ResNet-101     &      {24.0} / {27.3}      &    {14.2} / {18.0}        &  {4.2} /  {5.2}        \\ 	
   
   \rowcolor{mygray}	   {\textbf{SG2HOI+}$_{pool}$}   &  {ResNet-101}     &      { \textbf{24.3}} /  {\textbf{27.5}}      &    { \textbf{14.4}} /  {\textbf{18.2}}        &   {\textbf{4.3}} /   {\underline{5.3}}        \\ \bottomrule 
   
	\end{tabular}
}
\end{table}

\begin{table}[h]
	\centering
	\label{tab:vcoco}
	\caption{ Results on  V-COCO test set in terms of Scenario $1$ and  Scenario $2$.  Best result is marked with \textbf{bold} and the second best result
		is marked with \underline{underline}.  
	}
	\resizebox{\columnwidth}{!}{
		
		\begin{tabular}{llcll}
			\toprule
			
			& \textbf{Methods} & \multicolumn{1}{c}{\textbf{Backbone}} & \textbf{AP}$^{\#1}_{role}$ & \textbf{AP}$^{\#1}_{role}$ \\ \midrule
			\multirow{8}{*}{\textbf{Two-stage}} 
			&   InteractNet~\cite{gkioxari2018detecting}&ResNet-50-FPN &  40.0 & -    \\
			&   iCAN~\cite{gao2018ican}             &    ResNet-50     &  45.3 & 52.4    \\
			&   VSGNet~\cite{ulutan2020vsgnet}      &    ResNet-152    &  51.8 & 57.0   \\
			&   GPNN~\cite{qi2018learning}          &    Res-DCN-152   &  44.0 & -    \\ 
			&   PD-Net~\cite{zhong2020polysemy}     &   ResNet-50      &  52.6 & -    \\
			&   FCMNet~\cite{liu2020amplifying}     &   ResNet-50      &  53.1 & -    \\
			&   ACP~\cite{kim2020detecting}        &    ResNet-50      &  53.2 & -   \\
			&   IDN~\cite{li2020hoi}        &    ResNet-50      &  53.3 &  60.3  \\
			
			&   SG2HOI~\cite{he2021exploiting}      &     ResNet-50    &  53.3 &  59.4 \\
			\midrule
			
			\multirow{9}{*}{\textbf{One-stage}} 
			&  	UnionDet~\cite{kim2020uniondet}     &   ResNet-50-FPN  &    47.5 & 56.2      \\
			&   HOTR~\cite{kim2021hotr}             &  ResNet-50       &  55.2  &  64.4   \\
			&   AS-Net~\cite{chen2021reformulating} &  ResNet-50       &  53.9  &  -    \\
		
			&   HOI-Trans~\cite{zou2021end}         &  ResNet-50       &  52.9  &  - \\ 
		
			&   QPIC~\cite{tamura2021qpic}          &  ResNet-50       &  58.8  &  61.0   \\
			&   GEN-VLKT$_s$~\cite{liao2022gen}  &  ResNet-50  &   62.4 &  64.5 \\  
			
			&   SSRT~\cite{iftekhar2022look}			&ResNet-50       &  \textbf{63.7} & \textbf{65.9}  
			\\	
			
			&  \cellcolor{mygray}\textbf{SG2HOI+}$_{rand}$       	&  \cellcolor{mygray}ResNet-50       & \cellcolor{mygray}61.8 & \cellcolor{mygray}63.6  \\
			&  \cellcolor{mygray}\textbf{SG2HOI+}$_{clip}$    &  \cellcolor{mygray}ResNet-50       &  \cellcolor{mygray}{63.2} &  \cellcolor{mygray}{64.9}  \\

   &   {\cellcolor{mygray}\textbf{SG2HOI+}$_{pool}$}    &   {\cellcolor{mygray}ResNet-50}       &   {\cellcolor{mygray}\underline{63.6}} &   {\cellcolor{mygray}\underline{65.2}}  \\

			\bottomrule
	\end{tabular}}
\end{table}

\begin{table*}[]
	\centering
	\caption{Results comparison on the HICO-Det test set. We report three key techniques: backbone network, the datasets for object detection training and external knowledge for fair comparison.  Best result is marked with \textbf{bold} and the second best result
		is marked with \underline{underline}. $\dagger$ means the results are reproduced by the released codes.}
	\begin{tabular}{llccccccccc}
		\toprule
		& \multirow{3}{*}{\textbf{Methods}} & \multicolumn{1}{c}{\multirow{3}{*}{\textbf{Backbone}}} & \multirow{3}{*}{\textbf{Detector}} & \multirow{3}{*}{\textbf{Extra KG}} & \multicolumn{3}{c}{\textbf{Default}} & \multicolumn{3}{c}{\textbf{Known Object}} \\
		\multirow{17}{*}{\textbf{Two-stage}} &                          & \multicolumn{1}{c}{}                          &                           &                           & \textbf{Full}   & \textbf{Rare}   & \textbf{Non-Rare}  & \textbf{Full}     & \textbf{Rare}    & \textbf{Non-Rare}    \\ \midrule
		&  InteractNet~\cite{gkioxari2018detecting}   & ResNet-50-FPN  &  COCO     &  &   9.94   &  7.16     &  10.77       &     -      &     -     &     -               \\
		&  iCAN~\cite{gao2018ican}   &  ResNet-50 &  COCO &   &  14.84 & 10.45 & 16.15 & 16.26 & 11.33 & 17.73       \\
		&  VSGNet~\cite{ulutan2020vsgnet}   &  ResNet-152 &  COCO  & & 19.80 & 16.05 & 20.91    &    -  &  -  &      -   \\
		&  GPNN~\cite{qi2018learning}    & ResNet-101&   COCO  & & 13.11 & 9.34 & 14.23   &  -    & -   &   -       \\
		& TIN~\cite{li2020pastanet} &  ResNet-50  & COCO  & $\checkmark$  & 17.22  & 13.51  & 18.32 &  19.38 &  15.38 &  20.57 \\
		& CHG~\cite{wang2020contextual}  &   ResNet-50 &  COCO & &   17.57 &  16.85 &  17.78 &  21.00&   20.74&   21.08 \\
		&  PD-Net~\cite{zhong2020polysemy}   & ResNet-152 &  COCO  &  $\checkmark$ & 20.81 & 15.90 &22.28 &24.78 &18.88& 26.54   \\
		&  FCMNet~\cite{liu2020amplifying}    & ResNet-50  &   COCO  & $\checkmark$ &20.41 &17.34 &21.56& 22.04 &18.97& 23.12    \\
		&  ACP~\cite{kim2020detecting}    & ResNet-152 &  COCO   & $\checkmark$ & 20.59 & 15.92 & 21.98&    -   &   - &   -    \\
		&  IDN~\cite{li2020hoi}    & ResNet-50 & COCO   &     & 23.36  & 22.47 &  23.63  & 26.43  & 25.01 &  26.85     \\
		&  No-Fills~\cite{gupta2019no}  &     ResNet-152 &  COCO   & $\checkmark$  & 17.18  & 12.17  & 18.68   &  -  &   -     &    -    \\
		&  SCG~\cite{zhang2021spatially}   & ResNet-50-FPN&    COCO  & &  21.85 & 18.11&  22.97 &  -    & -   &   -       \\
		&  ATL~\cite{hou2021affordance} & ResNet-50  &  HICO-Det  & $\checkmark$ &23.81  & 17.43  & 25.72  & 27.38  & 22.09  & 28.96        \\
		
		&  SG2HOI~\cite{he2021exploiting}   & ResNet-50 &  COCO   & $\checkmark$ & 20.93  & 18.24  & 21.78  & 24.83 &  20.52 &  25.32     \\ \midrule
		
		\multirow{14}{*}{\textbf{One-stage}} 
		&\textbf{Two-decoder}: &&& &&&\\
		&     UnionDet~\cite{kim2020uniondet} & ResNet-50-FPN &     COCO  &   &  17.58  & 11.72  & 19.33 &  19.76 &  14.68  & 21.27 \\
		&HOTR~\cite{kim2021hotr}  & ResNet-50&   HICO-Det   &  &  25.10  & 17.34 &  27.42  & -  & -  & -\\
		& AS-Net~\cite{chen2021reformulating} & ResNet-50 &  HICO-Det  &  &  28.87  & 24.25 &  30.25 &  31.74 &  27.07 &  33.14 \\
		& GEN-VLKT$_{s}$ \cite{liao2022gen} &  ResNet-50 & HICO-Det   & $\checkmark$ & \textbf{33.75} &	{29.25} &	 {35.10}	 &\textbf{36.78}	 &\textbf{32.75} &	\textbf{37.99} \\  
		\cmidrule{2-11}
		
		&\textbf{Single-decoder}: & & & & & & & & & \\
		
		&  IP-Net~\cite{wang2020learning}  & Hourglass-104& COCO    &  &  19.56  & 12.79 &  21.58  & 22.05 &  15.77  &  23.92  \\
		& HOI-Trans~\cite{zou2021end} &  ResNet-50 &  HICO-Det      &    &   23.46   &  16.91   &  25.41  &   26.15   &  19.24  &   28.22  \\ 
		& QPIC~\cite{tamura2021qpic} & ResNet-50 &    HICO-Det  &  & 29.07 &  21.85  & 31.23  & 31.68  & 24.14  & 33.93 \\
		& SSRT~\cite{iftekhar2022look}  & ResNet-50&   - & $\checkmark$ & 30.36 & 25.42 & 31.83  & -  & - & - \\

		& \cellcolor{mygray}\textbf{SG2HOI}+$_{rand}$  & \cellcolor{mygray}ResNet-50 &   \cellcolor{mygray}HICO-Det & \cellcolor{mygray}$\checkmark$ & \cellcolor{mygray}31.54 &\cellcolor{mygray}26.73   & \cellcolor{mygray}32.41  & \cellcolor{mygray}32.52  & \cellcolor{mygray}28.62  & \cellcolor{mygray}34.18 \\
		
		& \cellcolor{mygray}\textbf{SG2HOI}+$_{clip}$  &\cellcolor{mygray}ResNet-50  & \cellcolor{mygray}HICO-Det &\cellcolor{mygray}$\checkmark$ &\cellcolor{mygray}{32.62} & \cellcolor{mygray}{28.54}  &  \cellcolor{mygray}{{35.17}} & \cellcolor{mygray}{35.47}  & \cellcolor{mygray}{31.36}  & \cellcolor{mygray}{36.02} \\

  	&  {\cellcolor{mygray}\textbf{SG2HOI}+$_{pool}$}  & {\cellcolor{mygray}ResNet-50} &  {\cellcolor{mygray}HICO-Det} &  {\cellcolor{mygray}$\checkmark$} &  {\cellcolor{mygray}\underline{33.14}} &  {\cellcolor{mygray}\textbf{29.27}}  &   {\cellcolor{mygray}{\textbf{35.72}}} &  {\cellcolor{mygray}\underline{35.73}}  &  {\cellcolor{mygray}\underline{32.01}}&  {\cellcolor{mygray}\underline{36.43}}\\
		\bottomrule      
	\end{tabular}
 \label{tab:hico}
\end{table*}
 \subsection{Results on HOI}
 { Following the mainstream HOI methods~\cite{zou2021end,kim2021hotr},  we first perform our model on the V-COCO dataset, and the results are shown in Table \ref{tab:vcoco} under two scenarios. From the results,  we could see our model not only significantly outperforms the two-stage models  but also achieves SoTA performance of the one-stage method. Specifically, in the two-stage category, SG2HOI$+{pool}$ outperforms those models like IDN by margins of $10.3$ and $4.9$ points under Scenario $1$ and Scenario $2$, respectively. It's worth highlighting that SG2HOI$+{pool}$ demonstrates a significant $17.8$\% average performance improvement compared to our previous work SG2HOI. Among the one-stage methods, while our model falls short of surpassing the competitive state-of-the-art model SSRT~\cite{iftekhar2022look}, it still secures the second-best position and notably outperforms other one-stage methods like HOTR and QPIC.}
 
 Table \ref{tab:hico} shows the results of the challenging dataset HICO-DET. For a fair comparison, we further report the utilization of  other important information that could benefit the model performance, such as  external knowledge (i.e., pose and language priors), object detection backbone, and training dataset for the object detection network. Regarding   one-stage methods, we further divide it into two subgroups, i.e., two- and single-decoder, according to whether or not to use two decoders for object and interaction prediction. From the results, we could make the following observations:  
 
  (1) The one-stage approach shows a significant and promising advantage over the two-stage models. For example, the best two-stage method ATL~\cite{hou2021affordance} lags behind both the best two-decoder based GEN-VLKT$_s$ and our  single-decoder based SG2HOI$+_{clip}$ by   $\sim$40\% and $\sim$36\% on average, respectively. In particular, when comparing our current model, SG2HOI$+_{pool}$, to our previous version, SG2HOI, a substantial improvement of approximately $50$\% is observed. This further confirms  that training both tasks concurrently within an end-to-end model yields noteworthy enhancements in performance.

   {
  (2) Our model does not surpass the top-performing model Gen-VLTK \cite{liao2022gen}, except within the Default Non-Rare setting, but our model (SG2HOI$+_{pool}$) achieves the second-best results across all HOI sub-tasks, outperforming other models by a significant margin. For instance, SG2HOI$+_{pool}$ surpasses the competitive SSRT \cite{iftekhar2022look} by a margin of about $2.5$ points, and this gap widens when compared to all two-step methods. With in-depth analysis, we think one reason is mainly because of the architectural distinction between our model and Gen-VLTK. More concretely, Gen-VLTK adopts a two-decoder architecture, employing separate decoders for object detection and interaction prediction, which allows for more refined presentation learning in both object and interaction classification tasks, as demonstrated in \cite{liao2022gen}. In contrast, our model employs a single-decoder architecture for the sake of ease of integrating SGG and HOI detection within a unified framework, but what our model sacrifices is the robust representation learning capabilities inherent to two-decoder architectures. Thus, moving forward, one potential avenue for enhancing our model is the exploration of a two-decoder architecture as the foundational structure. 
}
  

\subsection{Ablation Study}
In this subsection, we aim to evaluate the effectiveness of each component within our method through ablation experiments.  {These experiments are primarily conducted on the HICO-DET dataset. For each ablation model, we only remove the ablated part, but keep the other components at their best settings as our full model ($\textbf{SG2HOI+}_{pool}$), such as the same number of iterations, batch size, and hyperparameters, etc. Notably,  for all ablation results, we run each ablated model with five different random seeds and finally average them. 
} 

\subsubsection{Single task training} 
 {The central advantage of our method involves the joint training of SGG and HOI tasks to enhance both of them. We start by examining how the multi-task training strategy affects each individual task. Specifically, we first trained the SGG modules exclusively on the VG dataset. Once this training phase was completed, we froze those parameters and proceeded to train the HOI-related modules using the HICO-DET dataset.   The results of these separate training experiments are presented in Table \ref{tab:ab_st}.} Notably, we only report the primary results for SGDet in  SGG, while for HOI, we present results on the ``Default" setting.

 {Obviously, the separate training of each task on its respective dataset led to a relative degradation in their performance, although the extent of this decline differed between SGG and HOI. Specifically, the degradation observed in SGG was about $1$ point, which was smaller compared to the decline of approximately $3$ points observed in HOI. We attribute this difference to the sequential structure of our network. More specifically, the SGG module is positioned at the top of the network, while the HOI component is located at the bottom. Consequently, HOI detection relies more heavily on relation prediction, as SGs provide crucial cues for understanding interactions. On the other hand, the SGG module is less influenced when the HOI part is removed. This is because SGG predominantly relies on visual and holistic semantic-spatial features due to our hierarchical architecture, whereas the HOI task has a relatively lesser impact on SGG. }

\begin{table}[t]
	\centering
 
	\caption{Ablation study results on the single task training.}
	\resizebox{\columnwidth}{!}{
	\begin{tabular}{l|cccc|ccc}
		\toprule
		\multirow{3}{*}{\textbf{Models}} & \multicolumn{4}{c|}{\textbf{SGDet (VG)}}       & \multicolumn{3}{c}{\textbf{Default (HICO)}} \\
		& \textbf{R@50} & \textbf{R@100} & \textbf{mR@50} & \textbf{mR@100} & \textbf{Full} & \textbf{Rare} & \textbf{None-Rare} \\ \midrule
		SGG             				 & 24.6 &  28.4   & 10.5 & 12.9 & -     & -     & -     \\
		HOI            					 & -    &  -      & -    &  -   & 30.28 & 25.95 & 31.91 \\ \midrule
  \textsc{Full}                     & 25.9 &  30.3   & 11.4 & 13.9 & 33.14 & 29.27 & 35.72 \\ 
  \bottomrule
\end{tabular}
}
\label{tab:ab_st}
\end{table}

  \begin{table}[t]
	\centering
	 
	\caption{Ablation study results on the holistic spatial and semantic features.}
	\resizebox{\columnwidth}{!}{
	\begin{tabular}{l|cccc|ccc}
		\toprule
		\multirow{3}{*}{\textbf{Models}} & \multicolumn{4}{c|}{\textbf{SGDet (VG)}}       & \multicolumn{3}{c}{\textbf{Default (HICO)}} \\
		& \textbf{R@50} & \textbf{R@100} & \textbf{mR@50} & \textbf{mR@100} & \textbf{Full} & \textbf{Rare} & \textbf{None-Rare} \\ \midrule
		\textsc{-HSS}                    & 23.9 &  27.8   & 10.1  & 12.0 & 30.22 & 26.41 & 32.64 \\
		~\textsc{HSSd}                    & 24.2 &  28.2   & 10.7 & 12.5 & 30.54 & 26.49 & 33.38 \\
		~\textsc{HSSg}                    & 24.8 &  28.9   & 10.7 & 13.4 & 31.29 & 27.52 & 34.67 \\ \midrule
  \textsc{Full}                     & 25.9 &  30.3   & 11.4 & 13.9 & 33.14 & 29.27 & 35.72 \\  \bottomrule
  \end{tabular}
  }\label{tab:ab_hss}
  \end{table}

\begin{table}[t]
	\centering
 
	\caption{Ablation study results on the visual-language distillation.}
	\resizebox{\columnwidth}{!}{
	\begin{tabular}{l|cccc|ccc}
		\toprule
		\multirow{3}{*}{\textbf{Models}} & \multicolumn{4}{c|}{\textbf{SGDet (VG)}}       & \multicolumn{3}{c}{\textbf{Default (HICO)}} \\
		& \textbf{R@50} & \textbf{R@100} & \textbf{mR@50} & \textbf{mR@100} & \textbf{Full} & \textbf{Rare} & \textbf{None-Rare} \\ \midrule
	-\textsc{VL}$_{pool}$              & 24.3 &  27.9   & 10.3 & 12.3 & 31.49 & 27.34 & 34.21 \\
 ~\textsc{VL}$_{l_2}$              & 25.6 &  29.7   & 11.2 & 13.5 & 33.10 & 28.92 & 35.37 \\\midrule
  \textsc{Full}                     & 25.9 &  30.3   & 11.4 & 13.9 & 33.14 & 29.27 & 35.72 \\  
 \bottomrule
 \end{tabular}
 }\label{tab:ab_vl}
 \end{table}

 \begin{table}[t]
	\centering
 
	\caption{Ablation study results on the transformation of relation to interaction.}
	\resizebox{\columnwidth}{!}{
	\begin{tabular}{l|cccc|ccc}
		\toprule
		\multirow{3}{*}{\textbf{Models}} & \multicolumn{4}{c|}{\textbf{SGDet (VG)}}       & \multicolumn{3}{c}{\textbf{Default (HICO)}} \\
		& \textbf{R@50} & \textbf{R@100} & \textbf{mR@50} & \textbf{mR@100} & \textbf{Full} & \textbf{Rare} & \textbf{None-Rare} \\ \midrule
		 
		 -\textsc{FeTr}                   & 23.3 &  27.1   & 9.5 & 11.4 & 30.50 & 24.13 & 30.45 \\
		-\textsc{QuTr}                   & 24.9 &  28.5   & 10.9 & 13.0 & 33.37 & 28.24 & 34.60 \\
  -\textsc{R2ITr}   			     & 25.0 &  28.5   & 11.1 & 12.6 & 29.10 & 25.14 & 30.21 \\
		~\textsc{relQ}   			     & 22.6 &  26.1   & 10.4 & 11.3 & 29.29 & 24.03 & 30.62 \\ \midrule
		
	 \textsc{Full}                     & 25.9 &  30.3   & 11.4 & 13.9 & 33.14 & 29.27 & 35.72 \\  \bottomrule 
	\end{tabular}
}\label{tab:ab_r2i}
\end{table}

 \begin{table}[ht]
	\centering
	 
	\caption{Ablation  study results on data efficiency.}
	\begin{tabular}{l|ccc}
		\toprule
		\multirow{3}{*}{\textbf{Models}} &  \multicolumn{3}{c}{\textbf{Default (HICO)}} \\
		 & \textbf{Full} & \textbf{Rare} & \textbf{None-Rare} \\ \midrule
				\textsc{HOI}-Trans$^\star$      &         25.05 & 18.28 & 27.93 \\
		\textsc{QPIC}$^\star$             &        30.73 & 22.53 & 32.13 \\ \midrule
		
	 \textsc{Full}                 & 33.14 & 29.27 & 35.72 \\  \bottomrule 
	\end{tabular}
\label{tab:ab_data_eff}
\end{table}

\subsubsection{Holistic semantic-spatial feature} 
 {The importance of semantic and spatial features in SGG and HOI detection has been extensively proven in prior literature, particularly in the two-stage methods (e.g., \cite{zellers2018neural,tang2020unbiased,chen2019knowledge}). To assess the efficacy of these features our model learned, we conducted this ablation experiments on three variant models. Specifically, we first removed the $\mathrm{FCN}_{seg}$ module from our full model, denoted as \textsc{-HSS}, and retrained  \textsc{-HSS}  using the loss function detailed in Eq.(\ref{eq:loss_all}), wherein we set $\lambda_1$ to $0$, while retaining the remaining hyperparameters identical to our full model's configuration.}
Furthermore, we conducted an investigation into the influence of semantic embeddings by substituting the CLIP embeddings with Glove vectors~\cite{pennington2014glove} or discrete class labels, respectively denoted as \textsc{HSSg} and \textsc{HSSd}.  The results are presented in  Table \ref{tab:ab_hss}.

  {Generally,  the results  indicate a pronounced performance decline on the three ablated models. Notably, the most substantial degradation was observed in the \textsc{-HSS} ablation, exhibiting a decline of approximately $2\sim3$ points across both tasks. These findings underscore that semantic and spatial information have essential influence on  relation and interaction prediction. Besides, the embeddings from CLIP demonstrated greater efficacy compared to the other two embedding types (i.e., \textsc{HSSd} and \textsc{HSSg}), which experienced relatively minor degradation of less than $1$ point.
}

 
\subsubsection{Visual-linguistic knowledge distillation
}  {The distillation operation aims to align the encoded  features with the corresponding VL  feature from the pre-trained   CLIP, thereby mitigating the biased prediction in SGG and HOI datasets to some extent. In our best model, the alignment operation utilizes $\boldsymbol{f}^{en}_{\mathrm{[pool]}}$ to align with CLIP features. }
  {To valid the VL distillation module, we discard it from our model in this ablation study, denoted as  \textsc{-VL}$_{pool}$. Moreover, we investigate the influence of employing different alignment strategies in Eq. (\ref{eq:vl_align}). By  default, we use  $l_1$ to align both features, and further test  ${l_2}$ norm strategy denoted as \textsc{VL}$_{l_2}$. The results are presented in Table \ref{tab:ab_vl}.}
 
  {It is obvious that without the visual-linguistic knowledge alignment, the performance gains a relative decrease over two tasks, especially SGG on the unbiased metric of mR@K with $1.8$ point decline.  Also,  the degradation for HOI in the ``Rare'' category is  relatively larger. This is possibly because the VL model provides rich language prior knowledge for  encoded features and boosts the model's generalizability on those data-starving classes. Moreover, from the previous results of  ${\textbf{SG2HOI+}}_{cls}$ and $\textbf{SG2HOI+}_{pool}$, the latter consistently surpasses the former on the both SGG and HOI detection. We think that the main reason is the constraint way of condensing all features into one representation is much stronger than a single  token based strategy \texttt{[CLS]}, thereby resulting in better aligned features with CLIP features.}
 
   \begin{figure*}[t]
	\centering
	\includegraphics[width=1\linewidth]{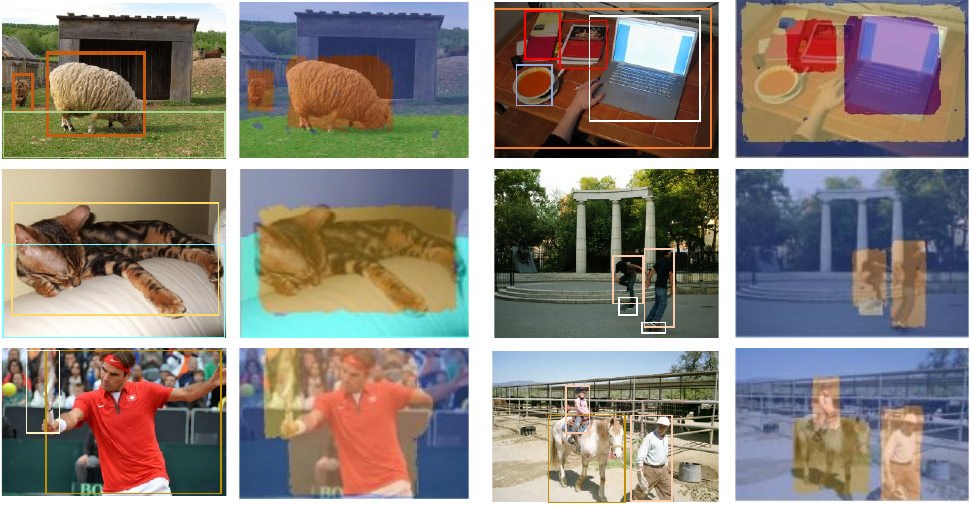}
	\caption{ The visualization results of our predicted bounding-box based segmentation. The left images are the input images from VG dataset and the corresponding ground-truth bounding boxes, while the right images are our predicted bounding-box based segmentations, serving as our holistic semantic-spatial representations. }
	\label{fig:vis_hss}
\end{figure*}
\subsubsection{Relation  to HOI  transformation} 
 {This experiment focuses on assessing the impact of two transformation strategies in our model: feature  and query transformation. For the former, we investigate a model variant denoted as $\textsc{-FeTr}$, which directly employs the feature $\boldsymbol{f}^{r2i}$ for HOI prediction, without the use of cross-attention mechanisms to integrate the feature $\boldsymbol{f}^{en}$ as in Eq. (\ref{eq:reltrans}). Also, we introduce a  variant model, termed \textsc{-R2ITr}, which directly takes as input the encoder image feature $\boldsymbol{f}^{en}$.  Concerning query transformation, we design the variant model that independently learns a set of HOI queries from scratch, referred to as $\textsc{-QuTr}$, without attending relation queries. Last, we test another strategy, denoted as $\textsc{relQ}$, where the relation queries  is used for the HOI queries. }

 {The results presented in Table \ref{tab:ab_r2i} demonstrate that these three models exhibit varying degrees of performance degradation. The observations from these results lead to the following conclusions: (1) directly predicting HOIs from the representation of an SG is not effective, as prediction errors from the SG branch could directly propagate to the HOI branch due to the one-on-one connection; (2) solely learning HOI queries from scratch is not effective, compared to further exploring the relation cues present in the relation query; (3) discarding the transformation module for SGs to HOIs results in disaster performance degradation on HOI detection; and (4) simply treating the relation query as the HOI query  is not viable, because of the intrinsic gap between the two tasks. }


\subsubsection{Data efficiency}
 {One main concern is that we integrate two tasks into one model and use more data to train our model. This potentially contributes to improving object detection performance,  due to the substantial overlap in object categories between the two datasets. In fact, more than $60$\% of objects in the HICO-DET dataset are shared with the VG dataset. This alignment between datasets undoubtedly enhances object detection capabilities. In order to assess whether the observed model enhancement can be attributed to the improved object detection performance, we conducted a comparative evaluation in which we chose QPIC \cite{tamura2021qpic} and HOI-Trans \cite{zou2021end} as the two representation models. Specifically, since both QPIC and HOI-Trans employ the pre-trained DETR on the MS-COCO dataset  for network initialization,  we first used the VG dataset further to fine-tune  the pre-trained DETR  and subsequently employed the refined DETR parameters to initialize QPIC and HOI-Trans. } The results are presented in Table \ref{tab:ab_data_eff}.

Obviously, we could see  an incremental improvement of approximately $1$ point on average happens over QPIC and HOI-Trans. Despite this enhancement, our model still surpasses both QPIC and HOI-Trans by a considerable margin, exhibiting a superiority of $3.6$ points on average compared to QPIC. This emphasizes that the improvements in our model are attributed not solely to the augmented dataset but also to the incorporation of other well-designed techniques.


\subsection{Visualization Results}
To qualitatively demonstrate the performance of SG2HOI+, we further visualize the predicted bounding-box based segmentation and some results of generated SGs and HOIs. 

Figure \ref{fig:vis_hss} shows $9$ cases of our predicted bounding-box based segmentation.   From the results, we could find that   our predicted bounding-box semantic segmentation are very close to the ground-truth bounding boxes and the segmentation are also rectangular as the bounding box.  However, there are also some failures. For instance, in the right image of the first row, our model fails to predict the smaller objects: a book and a plate. Also, our model fails to predict the skateboard for the right image in the second row. In conclusion, our model can segment those relatively bigger objects well rather than the small ones, but on the whole the embedded holistic semantic-spatial features  can still make a positive contribution to the downstream tasks SGG and HOI detection from the previous quantitative results. 

 \begin{figure*}[t]
	\centering
	\includegraphics[width=1\linewidth]{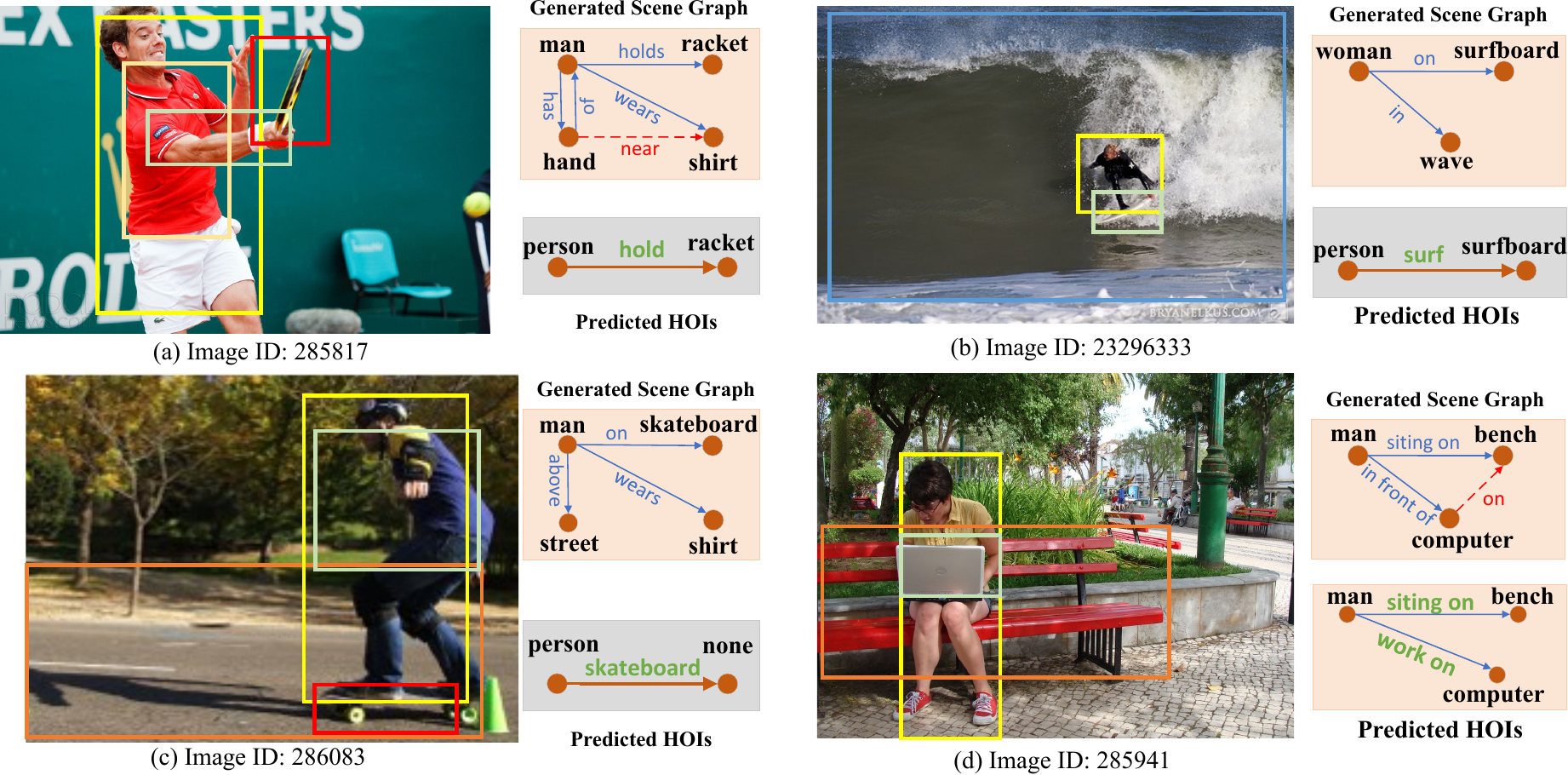}
	\caption{ The visualization results of our generated  scene graphs and human-object interactions in the VG dataset. Note that the red dashed line means the wrong prediction.   }
	\label{fig:vis_sg}
\end{figure*}

\begin{figure*}[t]
	\centering
	\includegraphics[width=1\linewidth]{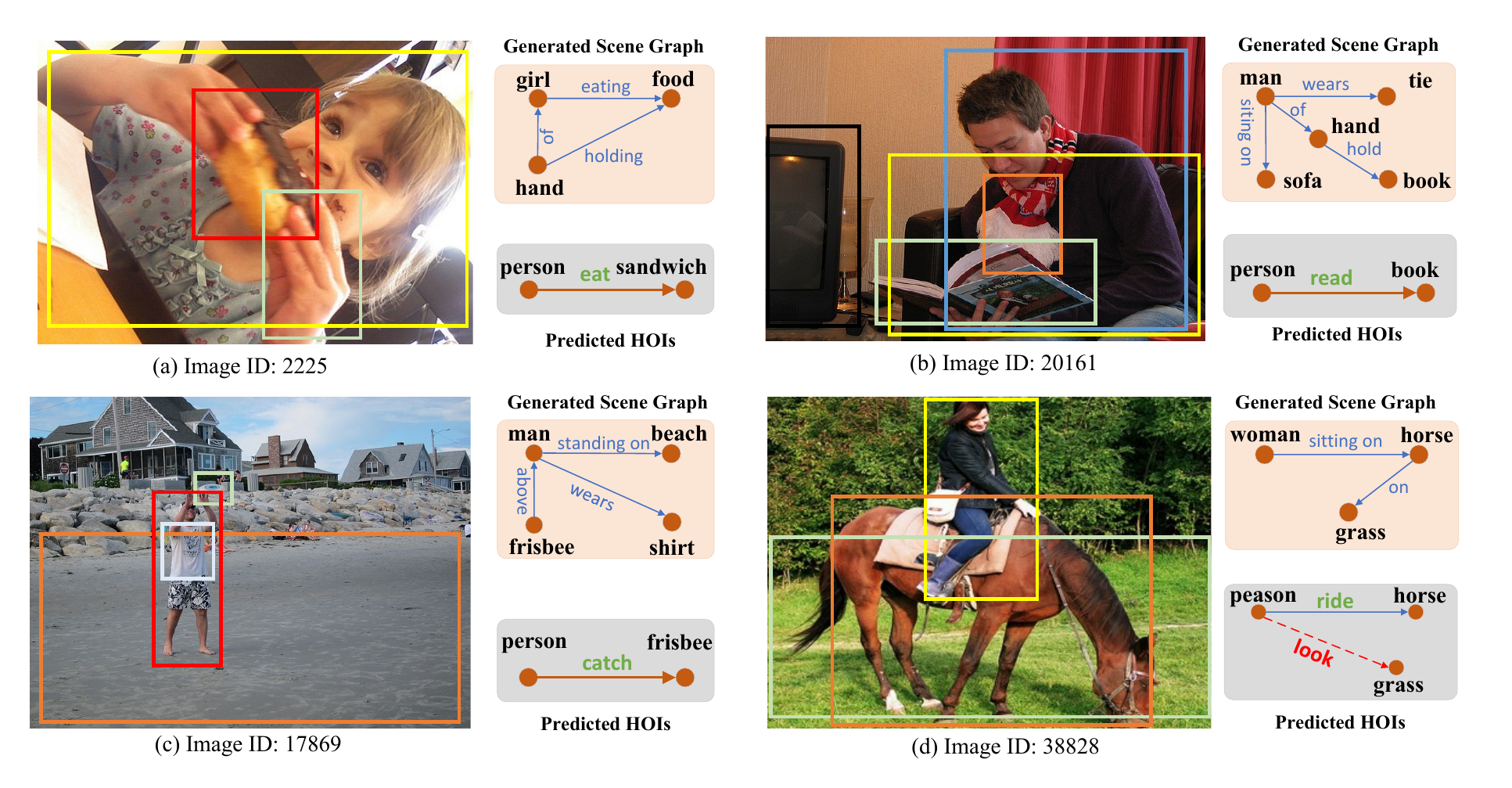}
	\caption{ The visualization results of our generated  scene graphs and human-object interactions in the V-COCO dataset. Note that the red dashed line means the wrong prediction.  }
	\label{fig:vis_hoi}
\end{figure*}

Figure \ref{fig:vis_sg} and \ref{fig:vis_hoi} provide visualized results showcasing SGs and HOIs produced by our {SG2HOI+} model for the VG and V-COCO datasets, respectively. The figures present input images along with their corresponding ground-truth bounding box annotations, and the SGs and HOIs generated by our end-to-end SG2HOI+ model in the right. An observation of the results reveals that our SG2HOI+ model effectively generates both SGs and HOIs for these datasets, albeit with some instances of failure. For example, in Figure \ref{fig:vis_sg}(a), where there is no  relationship between "hand" and ``shirt," our model predicts a ``near" predicate between them. Similarly, in Figure \ref{fig:vis_hoi}(d), our model overlooks the ``look" interaction, which highlights the need for capturing more nuanced features for smaller objects. Specifically, detecting the ``look" interaction necessitates capturing facial representations. However, due to limitations in our predicted bounding-box segmentation, our model struggles with accurate predictions for smaller objects.


\section{Conclusion}
 {Scene graph generation involves the task of detecting and localizing visual relationships among objects within an image, while human-object interaction detection is concerned with predicting interactions between humans and objects. In this study, we propose that visual relationships can offer vital contextual and intricate relational cues that substantially enhance the inference of human-object interactions. Building upon our prior work, SG2HOI, we present a unified end-to-end one-stage network that addresses both of these tasks. Concretely, we structure the two tasks sequentially, predicting human-object interactions from scene graphs. To achieve this, we first employ a relation transformer to generate scene graphs and subsequently introduce another transformer to predict interactions from a collection of visual relations. Furthermore, to effectively exploit semantic and spatial knowledge, we introduce a bounding-box-based segmentation technique to learn holistic semantic-spatial representations. Across a diverse array of benchmark datasets including Visual Genome, V-COCO, and HICO-Det, our model demonstrates superior performance compared to our previous work as well as several recent state-of-the-art methods in the domains of scene graph generation and human-object interaction detection.}

 {  \noindent\textbf{Limitations:} A primary limitation of our current model lies in its utilization of a single-decoder architecture, which may result in comparatively degraded feature learning capabilities for objects and predicates when contrasted with the two-decoder architecture models such as GEN-VLTK. }

 { In the future, one way for enhancing our model's capabilities involves the exploration of a two-branch architecture as the foundational structure. This strategic approach has the potential to yield better feature learning ability, not only concerning object and interaction detection but also extending to the task of scene graph generation.}

	{ 
		\bibliographystyle{IEEEtran}
		\bibliography{egbib}
	}

\end{document}